\pdfoutput=1

\documentclass[11pt]{article}

\usepackage{acl}

\usepackage{times}
\usepackage{latexsym}

\usepackage[T1]{fontenc}

\usepackage[utf8]{inputenc}
\usepackage{tipa}

\usepackage{microtype}
\usepackage{inconsolata}
\usepackage{linguex}
\usepackage{amsmath}
\usepackage{amssymb}
\usepackage{booktabs}
\usepackage{graphicx}
\usepackage{colortbl}
\usepackage{enumitem}
\usepackage{xcolor}
\usepackage{emoji}

\definecolor{cellgray}{HTML}{D3D3D3}

\newcommand{\jhu}{\textrm{\normalfont \textipa{C}}}
\newcommand{\yale}{\textrm{\normalfont \textipa{7}}}
\newcommand{\nyu}{\textrm{\normalfont \textipa{Z}}}

\title{Coloring the Blank Slate: Pre-training Imparts a Hierarchical\\Inductive Bias to Sequence-to-sequence Models}

\author{Aaron Mueller$^\jhu{}$~\;~ Robert Frank$^\yale{}$~\;~ Tal Linzen$^\nyu{}$ \\ \bf{Luheng Wang}$^\nyu{}$~\;~ \bf{Sebastian Schuster}$^\nyu{}$ \\
  $^\jhu{}$Johns Hopkins University~\;~\;~  $^\yale{}$Yale University~\;~\;~ $^\nyu{}$New York University \\
  \texttt{amueller@jhu.edu, schuster@nyu.edu} \\}

\begin{document}
\maketitle
\begin{abstract}
Relations between words are governed by hierarchical structure rather than linear ordering. Sequence-to-sequence (seq2seq) models, despite their success in downstream NLP applications, often fail to generalize in a hierarchy-sensitive manner when performing syntactic transformations---for example, transforming declarative sentences into questions. However, syntactic evaluations of seq2seq models have only observed models that were \emph{not} pre-trained on natural language data before being trained to perform syntactic transformations, in spite of the fact that pre-training has been found to induce hierarchical linguistic generalizations in language models; in other words, the syntactic capabilities of seq2seq models may have been greatly understated. We address this gap using the pre-trained seq2seq models T5 and BART, as well as their multilingual variants mT5 and mBART. We evaluate whether they generalize hierarchically on two transformations in two languages: question formation and passivization in English and German. We find that pre-trained seq2seq models generalize hierarchically when performing syntactic transformations, whereas models trained from scratch on syntactic transformations do not. This result presents evidence for the learnability of hierarchical syntactic information from non-annotated natural language text while also demonstrating that seq2seq models are capable of syntactic generalization, though only after exposure to much more language data than human learners receive.
\end{abstract}

\setlength{\Exlabelwidth}{0.25em}
\setlength{\SubExleftmargin}{1.35em}
\setlength{\Extopsep}{0.5\baselineskip}

\section{Introduction}
\begin{figure}[!ht]
    \centering
    \includegraphics[width=\linewidth]{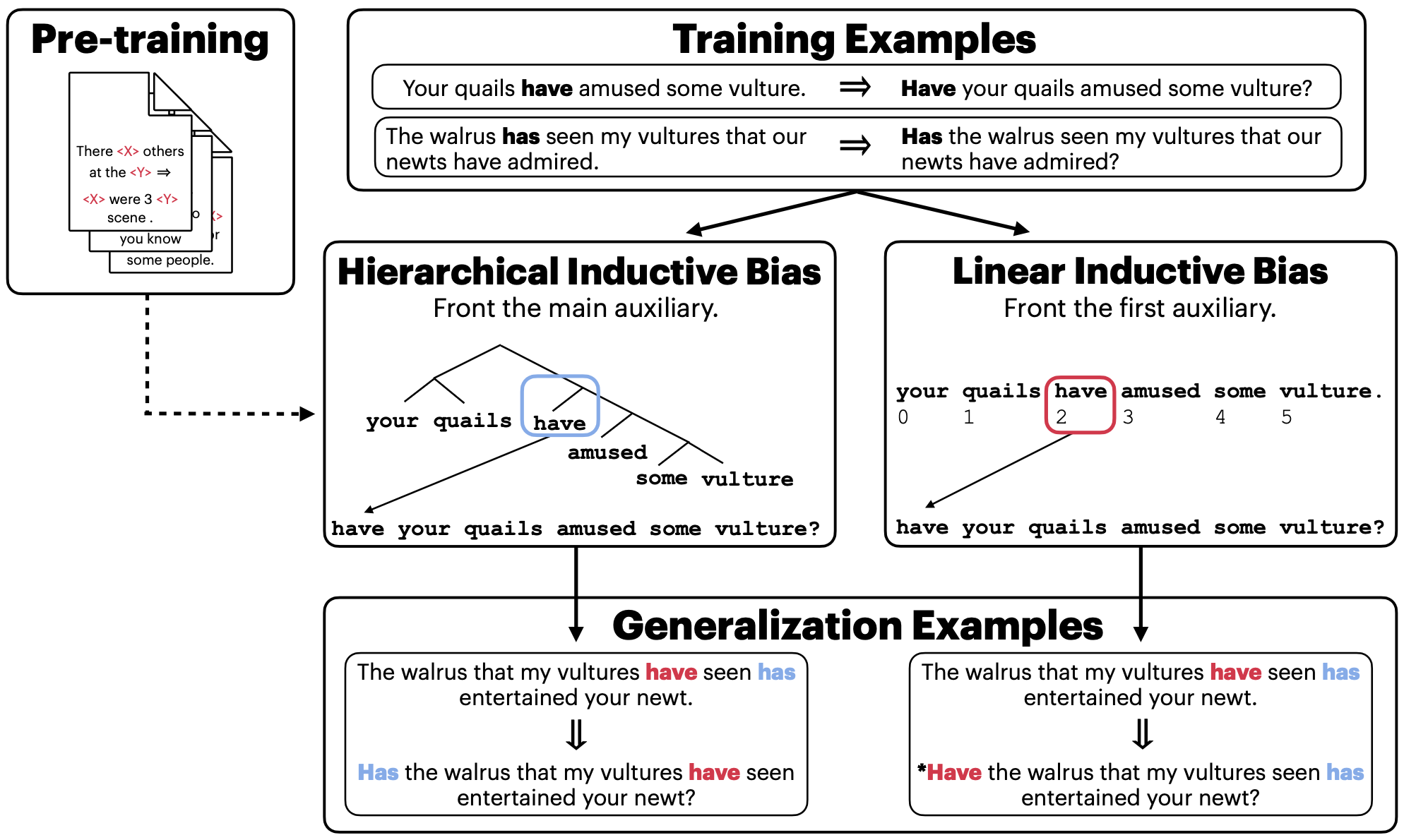}
    \caption{The poverty of the stimulus experimental design. We fine-tune pre-trained seq2seq models and train small seq2seq models from scratch to perform syntactic transformations. The training set contains ambiguous examples consistent with hierarchical and linear transformation rules. The generalization set contains examples where only the hierarchical rule results in the correct output. Pre-trained models generalize using the hierarchical rule, while models trained from scratch generalize using the linear rule.}
    \label{fig:povstim}
\end{figure}

Human language is structured hierarchically. In NLP tasks like natural language inference, syntactic competence is a prerequisite for robust generalization \citep[e.g.,][]{mccoy2019hans}. Probing studies have found that masked language models (MLMs) contain hierarchical representations \citep{tenney2019bert,hewitt2019structural,clark-etal-2019-bert}, while behavioral studies of recurrent neural language models \citep{linzen2016assessing,marvin2018targeted,wilcox2018rnn,vanschijndel2019quantity} and MLMs \citep{goldberg2019bert,hu2020systematic} have found that models are largely able to capture long-range syntactic dependencies that require hierarchical representations of sentences.

Recent evidence suggests that MLMs like BERT \citep{devlin2019bert} and RoBERTa \citep{liu2019roberta} can learn to make hierarchical linguistic generalizations through exposure to text \citep{warstadt2020linguistic}, though acquiring many of these linguistic generalizations requires large amounts of data \citep{warstadt2020learning}.  However, this evidence comes from binary acceptability judgment tasks, where a classifier head is attached to an MLM and the model is fine-tuned to classify which sentence in a given minimal pair is consistent with a hierarchical linguistic generalization, rather than a positional surface heuristic. Consider the following two transformations of Example~\ref{ex:bias}:

\ex.\label{ex:bias}The yak that your unicorns \textbf{\textcolor{red}{have}} amused \textbf{\textcolor{blue}{hasn't}} entertained a newt.
    \a.\label{ex:bias_correct} \textbf{\textcolor{blue}{Hasn't}} the yak that your unicorns \textbf{\textcolor{red}{have}} amused entertained a newt?
    \b.\label{ex:bias_incorrect} *\textbf{\textcolor{red}{Have}} the yak that your unicorns amused hasn't entertained a newt?

Example~\ref{ex:bias_correct} correctly forms the question by moving the main auxiliary verb to the front of the sentence, while \ref{ex:bias_incorrect} relies on the incorrect positional heuristic that the first auxiliary in the declarative sentence should be moved to the front of the sentence. When differentiating grammatical and ungrammatical auxiliary movements, a model could rely on distributional information \citep{lewis2001learnability} such as bigram heuristics \citep{reali2005uncovering,kam2008bigrams} to make correct judgments in many cases, so high performance on binary classification tasks may overstate the syntactic competence of a model.

By contrast, \emph{performing} a syntactic trans\-formation---e.g., given a declarative sentence like Example~\ref{ex:bias} as input, transforming it into a polar question like \ref{ex:bias_correct}---is more difficult. It requires multiple complex but systematic operations that rely on hierarchical structure, including movement, number agreement, and---in languages that have grammatical case, such as German---case reinflection. Evaluations of syntactic transformational abilities can therefore act as more targeted behavioral indicators of syntactic structural representations in neural models. \citet{mccoy2018poverty} evaluate non-pre-trained recurrent sequence-to-sequence (seq2seq) models \citep{sutskever2014sequence} on the question formation task, finding that they rely on linear/positional surface heuristics rather than hierarchical structure to perform this syntactic transformation. More recent studies have also exclusively considered recurrent seq2seq models and Transformer models \citep{petty2021transformers} trained from scratch on other transformations like tense reinflection \citep{mccoy2020trees} and passivization \citep{mulligan2021structure}, finding similar results. These studies were designed to understand the inductive biases of various seq2seq architectures, which is why they do not pre-train the models on non-annotated natural language data before training them to perform syntactic transformations.

\begin{table*}[]
    \centering
    \resizebox{\linewidth}{!}{
    \begin{tabular}{lp{10cm}p{11cm}}
    & \fcolorbox{black}{white}{\phantom{\vrule width 1.25mm height 1.25mm}} Train, dev, test & \fcolorbox{black}{cellgray}{\phantom{\vrule width 1.25mm height 1.25mm}} Generalization \\
    \toprule
    \textbf{Structure} & Question Formation & Passivization \\\midrule
    No RC/PP & quest: some xylophones have remembered my yak. \newline$\rightarrow$ have some xylophones remembered my yak? & passiv: your quails amused some vulture. \newline$\rightarrow$ some vulture was amused by your quails. \\\midrule
    RC/PP on object & quest: my zebras have amused some walrus who has waited. \newline$\rightarrow$ have my zebras amused some walrus who has waited? & passiv: some tyrannosaurus entertained your quail behind your newt. \newline$\rightarrow$ your quail behind your newt was entertained by some tyrannosaurus. \\\midrule
    RC/PP on subject & \cellcolor{cellgray} quest: my vultures that our peacock hasn't applauded haven't read. \newline$\rightarrow$ haven't my vultures that our peacock hasn't applauded read? & \cellcolor{cellgray} passiv: the zebra upon the yak confused your orangutans. \newline$\rightarrow$ your orangutans were confused by the zebra upon the yak. \\
    \bottomrule
    \end{tabular}}
    \caption{The distribution of syntactic structures in the train, test, and generalization sets. To expose the model to all structures during training and fine-tuning, we also include identity transformations for all structures using the ``decl:'' prefix, where the input and output sequences are the same declarative or active sentence (see \S\ref{sec:data}). We use the test set to evaluate whether models have learned the task on in-distribution examples, and the generalization set to evaluate whether models generalize hierarchically. See Appendix~\ref{app:train_test_gen_de} for example sentences in German.}
    \label{tab:train_test_gen}
\end{table*}

In this study, we create German datasets and modify English datasets for evaluating the inductive biases of pre-trained models. We use these datasets to analyze performance in monolingual and zero-shot cross-lingual settings. Further, we analyze \emph{how} pre-trained models perform syntactic transformations. Our findings indicate that pre-trained models generally perform syntactic transformations in a hierarchy-sensitive manner, while non-pre-trained models (including randomized-weight versions of pre-trained models) rely primarily on linear/positional heuristics to perform the transformations. This finding presents additional evidence to \citet{warstadt2020learning} and \citet{warstadt2020linguistic} for the learnability of hierarchical syntactic information from natural language text input. Our code and data are publicly available.\footnote{\url{https://github.com/sebschu/multilingual-transformations}}

\section{Syntactic Transformations}
\label{sec:transformations}
\subsection{Languages}
We evaluate on syntactic transformations in English and German. We choose English to allow for comparisons to previous results \citep{mccoy2018poverty,mulligan2021structure}. We further extend our evaluations to German because it exhibits explicit case marking on determiners and nouns; this typological feature has been found to increase the sensitivity of language models to syntactic structure \citep{ravfogel2019synthetic}. This allows us to compare transformational abilities for languages with different levels of surface cues for hierarchy.

\subsection{Tasks}
We employ a \emph{poverty of the stimulus} experimental design \citep{wilson2006learning}, where we train the model on examples of a linguistic transformation that are compatible with either a hierarchical rule or a linear/positional rule, and then evaluate the model on sentences where only the hierarchical rule leads to the generalization pattern that is consistent with the grammar of the language (Figure~\ref{fig:povstim}).\footnote{There are other rules that could properly transform the stimuli we use, but we find that the models we test do learn one of these rules or the other.} In other words, we are interested in whether T5 and mT5 (henceforth, (m)T5), as well as BART and mBART (henceforth, (m)BART), demonstrate a \textbf{hierarchical inductive bias},\footnote{When multiple generalizations are consistent with the training data, ``inductive bias'' refers to a model's choice of one generalization over others.} unlike the linear inductive bias displayed in prior work by non-pre-trained models.

We focus on two syntactic transformation tasks: \textbf{question formation} and \textbf{passivization}. See Table~\ref{tab:train_test_gen} for a breakdown of which structures we present to the model during training and which we hold out to evaluate hierarchical generalization. See Table~\ref{tab:data_examples} for examples of hierarchical and linear generalizations for each transformation.

\paragraph{Question formation.} In this task, a declarative sentence is transformed into a polar question by moving the main (matrix) auxiliary verb to the start of the sentence; this hierarchical rule is called \textsc{move-main}. The linear rule, \textsc{move-first}, entails moving the linearly first auxiliary verb to the front of the sentence. Examples of both rules are provided in Figure~\ref{fig:povstim} and Example~\ref{ex:bias}. We train the model on sentences with no relative clauses (RCs) or with RCs on the object, where the first auxiliary verb is always the matrix verb. Disambiguating examples are those which place RCs on the subject, where the matrix auxiliary verb is the linearly second auxiliary in the sentence.

\begin{table*}[]
    \centering
    \small
    \resizebox{0.9\linewidth}{!}{
    \begin{tabular}{p{5cm}p{5.5cm}p{5cm}}
    \toprule
    Input & Output (hierarchical) & Output (linear) \\
    \midrule
    quest: My unicorn that \textbf{\textcolor{red}{hasn't}} amused the yaks \textbf{\textcolor{blue}{has}} eaten. & \textbf{\textcolor{blue}{Has}} my unicorn that hasn't amused the yaks eaten? & \textbf{\textcolor{red}{Hasn't}} my unicorn that amused the yaks has eaten? \\\midrule
    quest: Die Hunde, die deine Löwen bewundern \textbf{\textcolor{red}{können}}, \textbf{\textcolor{blue}{haben}} gewartet. & \textbf{\textcolor{blue}{Haben}} die Hunde, die deine Löwen bewundern können, gewartet? & \textbf{\textcolor{red}{Können}} die Hunde, die deine Löwen bewundern, haben gewartet? \\
    \midrule\midrule
    passiv: Her walruses above \textbf{\textcolor{red}{my unicorns}} annoyed \textbf{\textcolor{blue}{her quail}}. & \textbf{\textcolor{blue}{Her quail}} was annoyed by her walruses above my unicorns. & \textbf{\textcolor{red}{My unicorns}} were annoyed by her walruses. \\\midrule
    passiv: Unsere Papageie bei \textbf{\textcolor{red}{meinen Dinosauriern}} bedauerten \textbf{\textcolor{blue}{unsere Esel}}. & \textbf{\textcolor{blue}{Unsere Esel}} wurden von unseren Papageien bei meinen Dinosauriern bedauert. & \textbf{\textcolor{red}{Meine Dinosaurier}} wurden von unseren Papageien bedauert.\\
    \bottomrule
    \end{tabular}}
    \caption{Examples from the generalization set with hierarchical- and linear-rule transformations. Glossed German examples are provided in Appendix~\ref{app:train_test_gen_de}.}
    \label{tab:data_examples}
\end{table*}

In English, we use the auxiliaries ``has'', ``hasn't'', ``have'', and ``haven't'', with past participle main verbs (e.g., ``have \emph{entertained}'', ``has \emph{amused}''). We use affirmative and negative forms to distinguish between the multiple auxiliaries: exactly one of the auxiliaries in such sentences is negative and the other is positive (counterbalanced across examples). As a result, we can determine whether the induced mapping is linear or hierarchical. In German, negation is realized as a separate word that is not fronted with the auxiliary. To distinguish the multiple auxiliaries, we therefore use the modal ``können'' (\emph{can}) along with the auxiliary ``haben'' (\emph{have}), together with infinitival or past participle main verbs as appropriate. This allows us to distinguish models with a hierarchical bias from those with a linear bias on the basis of the fronted auxiliary.

\paragraph{Passivization.} In this task, an active sentence is transformed into a passive sentence by moving the object noun phrase (NP) to the front of the sentence (\textsc{move-object}). Our training examples are also compatible with a linear rule, \textsc{move-second}, in which the linearly second NP moves to the front of the sentence. We train on sentences with no prepositional phrases (PPs) or with PPs modifying the object, where the second NP is always the object. Disambiguating examples are those which place prepositional phrases (PPs) on the subject, where the object is the linearly third NP in the sentence.

Passivization additionally requires other movements, insertions, tense reinflection, and (for German) case reinflection. In Examples~\ref{ex:passiv_en} and \ref{ex:passiv_de} below, the object (in \textcolor{blue}{blue}) is fronted; `be'/`werden' (in \textcolor{red}{red}) is inserted and inflected to agree with the fronted NP; the original subject NP (in \textcolor{brown}{brown}) is moved to a `by'/`von' phrase after the inserted verb; and the main verb (in \textcolor{orange}{orange}) is reinflected to be a past participle or infinitive. In German, the case of the NPs (reflected largely in the determiners) must be reinflected, and the main verb needs to be moved to the end of the sentence.

\ex.\textit{English Passivization}:\label{ex:passiv_en}
    \a. \textcolor{brown}{Your quails} \textcolor{orange}{amused} \textcolor{blue}{some vulture}.
    \b. \textcolor{blue}{Some vulture} \textcolor{red}{was} \textcolor{orange}{amused} by \textcolor{brown}{your quails}.
    
\ex.\textit{German Passivization}:\label{ex:passiv_de}
    \ag. \textcolor{brown}{Ihr} \textcolor{brown}{Esel} \textcolor{orange}{unterhielt} \textcolor{blue}{meinen} \textcolor{blue}{Salamander}.\\
        Your.\textsc{nom} donkey entertained my.\textsc{acc} salamander.\\
    \bg. \textcolor{blue}{Mein} \textcolor{blue}{Salamander} \textcolor{red}{wurde} von \textcolor{brown}{ihrem} \textcolor{brown}{Esel} \textcolor{orange}{unterhalten}.\\
        My.\textsc{nom} salamander was from your.\textsc{dat} donkey entertained.\\

\section{Experimental Setup}
\subsection{Data}\label{sec:data}
We modify and supplement the context-free grammar of \citet{mccoy2020trees} to generate our training and evaluation data.\footnote{We generate our evaluation set such that it consists of grammatical but semantically improbable sentences which are unlikely to occur in a natural language corpus. This is to alleviate the confound of token collocations in the pre-training corpus.} For each transformation, our training data consists of 100,000 examples with an approximately 50/50 split between identity examples (where the input and output sequences are the same) and transformed examples. The identity examples include the full range of declarative or active structures (including sentences with RCs/PPs on subjects), thereby exposing the network to the full range of input structures we test. For the transformed examples, however, training data includes only examples with no RCs/PPs or RCs/PPs on the object NP---i.e., cases that are compatible with both the hierarchical and linear rules. We also generate development and test sets consisting of 1,000 and 10,000 examples, respectively, containing sentences with structures like those used in training; these are for evaluating in-distribution transformations on unseen sentences.

For each transformation, we also generate a generalization set consisting of 10,000 transformed examples with RCs/PPs on the subject NP. For such examples, models relying on the linear rules will not generalize correctly.

\subsection{Models}
We experiment with T5 \citep{raffel2020t5} and BART \citep{liu2020mbart}, two English pre-trained sequence-to-sequence models. We also experiment with their multilingual variants mT5 \citep{xue2021mt5} and mBART \citep{liu2020mbart}.\footnote{We use HuggingFace implementations \citep{wolf2020huggingface}.} These are 12-layer Transformer-based \citep{vaswani2017transformer} architectures with bidirectional encoders and autoregressive decoders. While we use the base sizes of (m)T5, we use the large sizes of (m)BART to keep the sizes of the models similar.

When fine-tuning (m)T5 and (m)BART, we use task prefixes in the source sequence. We use ``quest:'' for question formation and ``passiv:'' for passivization. As in previous work, we also include identity transformation examples (prefixed with ``decl:''), i.e., examples for which the model has to output the unchanged declarative or active sentence. When training seq2seq baselines from scratch, we follow \citet{mccoy2020trees} and append the task markers to the end of the input sequence. 

For fine-tuning on syntactic transformations, we use batch size $128$ and initial learning rate $5\times 10^{-5}$. We fine-tune for $10$ epochs and evaluate every $500$ iterations. We find that the validation loss generally converges within 1--2 epochs.

To confirm the finding of \citet{mccoy2020trees} and \citet{petty2021transformers} that non-pre-trained models fail to generalize hierarchically, we also train baseline seq2seq models similar to the models used in those studies. We implement 1- and 2-layer LSTM-based seq2seq models, as well as 1- and 2-layer Transformer-based seq2seq models where the Transformers have 4 attention heads.\footnote{Our implementations are based on the syntactic-transformation-focused \texttt{transductions} repository: \url{https://github.com/clay-lab/transductions}} We find that the 1-layer models consistently achieve higher sequence accuracies on the dev sets, so we focus on the 1-layer baselines. We re-use all hyperparameters from \citet{mccoy2020trees}. 
All baseline scores are averaged over 10 runs.

\subsection{Metrics}
For all transformations, we are primarily interested in \emph{sequence accuracy}: is each token in the target sequence present in the proper order in the predicted sequence? However, it is possible that models could generalize hierarchically while making some other mistake, so we also use two more relaxed metrics. For question formation, we use \emph{main auxiliary accuracy}, which evaluates whether the correct auxiliary was moved to the front of the sentence. The first word in the target sequence is always the main auxiliary verb, so we calculate main auxiliary accuracy by checking if the first word is the same in the predicted and target sequences. For passivization, we use \emph{object noun accuracy}, which measures whether the correct object noun was moved to the subject position. The second word in the target sequence is always the original object noun, so we calculate object noun accuracy by checking if the second word is the same in the predicted and target sequences.

\section{Results}\label{sec:results}
\begin{table}[]
    \centering
    \resizebox{\linewidth}{!}{
    \begin{tabular}{lrrrr}
    \toprule
    & \multicolumn{2}{c}{\textbf{Question Formation}} & \multicolumn{2}{c}{\textbf{Passivization}}\\\cmidrule(lr){2-3}\cmidrule(lr){4-5}
    \textbf{Model} & English & German & English & German \\\midrule
    LSTM & 0.95 & 0.94 & 0.97 & 0.97 \\
    Transformer & 0.95 & 0.93 & 0.98 & 0.98 \\
    \midrule
    T5 & 1.00 & -- & 1.00 & -- \\
    mT5 & 1.00 & 1.00 & 1.00 & 1.00 \\
    BART & 0.96 & -- & 0.95 & -- \\
    mBART & 1.00 & 1.00 & 1.00 & 1.00 \\
    \bottomrule
    \end{tabular}}
    \caption{Sequence accuracies on the (in-distribution) test sets for English and German syntactic transformations. All models learn the in-distribution transformations.}
    \label{tab:indomain_seq}
\end{table}

\begin{table}[]
    \centering
    \resizebox{\linewidth}{!}{
    \begin{tabular}{lrrrr}
    \toprule
    & \multicolumn{2}{c}{\textbf{Question Formation}} & \multicolumn{2}{c}{\textbf{Passivization}}\\\cmidrule(lr){2-3}\cmidrule(lr){4-5}
    \textbf{Model} & English & German & English & German \\\midrule
    LSTM & 0.11 & 0.33 & 0.05 & 0.44 \\
    Transformer & 0.07 & 0.05 & 0.04 & 0.07 \\
    \midrule
    T5 & 0.87 & -- & 1.00 & -- \\
    mT5 & 0.99 & 1.00 & 1.00 & 1.00 \\
    BART & 0.96 & -- & 1.00 & -- \\
    mBART & 0.59 & 0.82 & 0.80 & 0.98 \\
    \bottomrule
    \end{tabular}}
    \caption{Main auxiliary accuracies (for question formation) or object noun accuracies (for passivization) on the generalization sets for English and German syntactic transformations. Only pre-trained models generalize hierarchically.}
    \label{tab:gen_firstsecond}
\end{table}

\paragraph{All models learn the in-distribution transformations.} We first present results on unseen sentences whose structures were seen in training, where both the hierarchical and the linear rules result in correct generalization (Table~\ref{tab:indomain_seq}). All models perform well in this setting, including the LSTM- and Transformer-based models trained from scratch. However, (m)T5 converges to higher sequence accuracies than the non-pre-trained models. Additionally, while the non-pre-trained models require about 15--20 epochs of training to converge to a high score, (m)T5 and (m)BART converge to near-perfect sequence accuracy after only a fraction of an epoch of fine-tuning.

\begin{figure*}
    \centering
    \includegraphics[height=3.5cm]{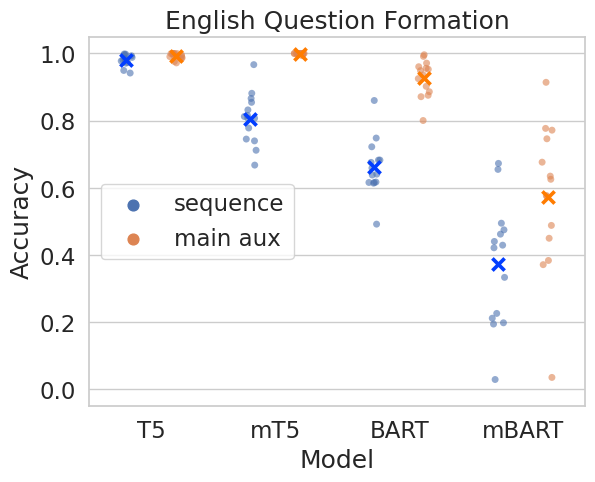}
    \hfill
    \includegraphics[height=3.5cm,trim={2cm 0 0 0},clip]{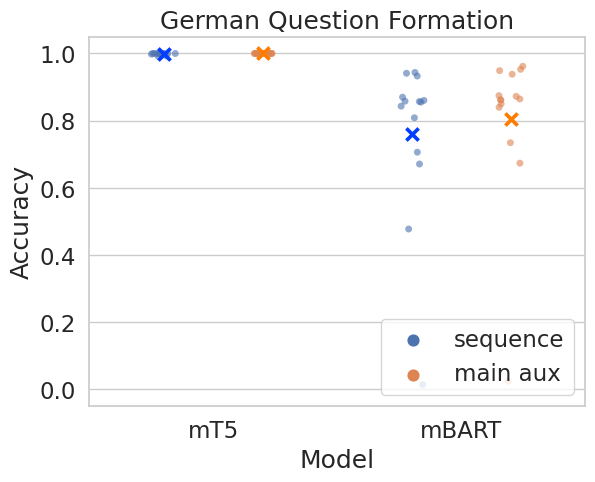}
    \hfill
    \includegraphics[height=3.5cm,trim={2cm 0 0 0},clip]{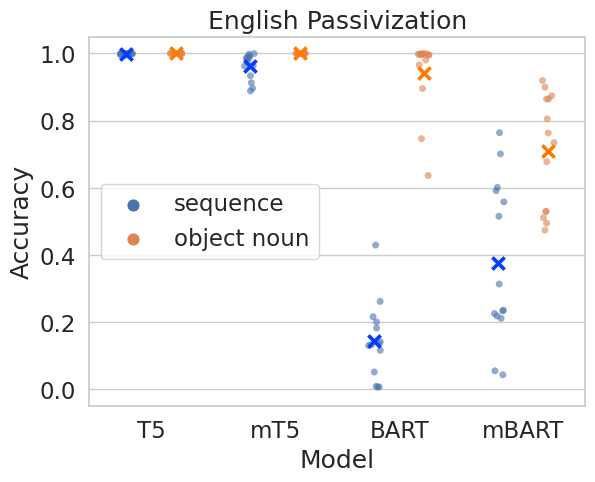}
    \hfill
    \includegraphics[height=3.5cm,trim={2cm 0 0 0},clip]{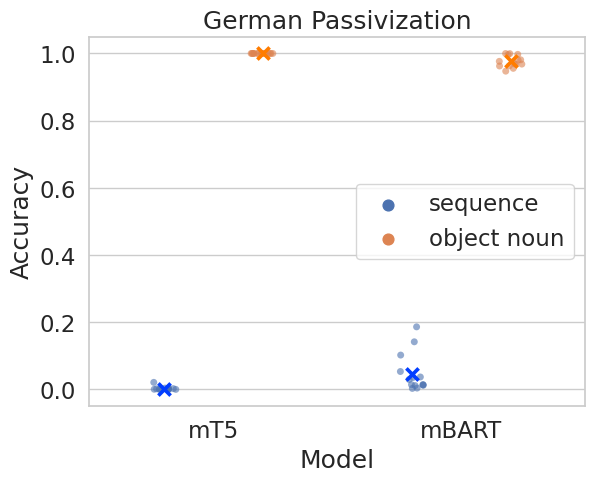}
    \caption{Accuracies at every 500 fine-tuning iterations across 10 epochs of fine-tuning on each syntactic transformation. Xs indicate mean accuracies across epochs. T5 models are generally better at performing syntactic transformations than BART models. Monolingual models tend to achieve higher accuracies than multilingual models. We present full learning curves in Appendix~\ref{app:learning_curves}.}
    \label{fig:all_models_scatter}
\end{figure*}

\paragraph{Only pre-trained models generalize hierarchically.} Evaluations on the generalization-set examples (where the linear rule leads to incorrect generalization) reveal that none of the trained-from-scratch models have learned the hierarchical rule. These models consistently stay at or near 0\% sequence accuracy on the generalization set throughout training, so we present main auxiliary/object noun accuracies (Table~\ref{tab:gen_firstsecond}). Accuracy remains low even on these more forgiving metrics, indicating that the non-pre-trained models have not acquired the hierarchical rules.

Low accuracies do not necessarily indicate reliance on the linear \textsc{move-first} or \textsc{move-second} rules. To test whether the non-pre-trained models have learned the linear rules, we implement metrics which calculate the proportion of generalization-set examples for which the \textsc{move-first} rule (for question formation) or \textsc{move-second} rule (for passivization) were used; we refer to these as the move-first frequency and move-second frequency, respectively. For each model and language, the sum of the main auxiliary accuracy and move-first frequency for question formation is $\approx1.0$; the sum of the object noun accuracy and move-second frequency for passivization is also $\approx1.0$. Thus, where the model did not move the main auxiliary or object noun, it generally used the linear rule. In other words, \textbf{the non-pre-trained models demonstrate \emph{linear} inductive biases.} This finding is in line with prior evaluations of non-pre-trained seq2seq models \citep{mccoy2020trees,mulligan2021structure,petty2021transformers}.\footnote{Nonetheless, higher accuracies on German transformations support the hypothesis that more explicit cues to syntactic structure (here, case-marked articles and nouns) allow models to learn hierarchical syntactic generalizations more easily. This agrees with the findings of \citet{ravfogel2019synthetic} and \citet{mueller2020multiling}.}

By contrast, (m)T5 and (m)BART achieve very high main auxiliary/object noun accuracies on the generalization sets. mBART struggles with English question formation, achieving an average 59\% main auxiliary accuracy throughout fine-tuning. However, it does achieve a maximum accuracy $>$90\%, indicating that it is capable of hierarchical generalization after observing certain training examples. These accuracies are still well above the $\approx$0\% accuracies of the non-pre-trained models.

Because sequence accuracy on the generalization set is often unstable for all pre-trained models, we present plots showing the distribution of accuracies sampled at every 500 fine-tuning iterations throughout 10 epochs of fine-tuning (Figure~\ref{fig:all_models_scatter}). Each pre-trained model learns the in-distribution transformation before the first 500 iterations of fine-tuning, so each plotted accuracy can be taken as indicative of model preferences after they have learned the transformations. (m)T5's sequence accuracies are generally close to 100\% for all transformations except German passivization; this is far better than the non-pre-trained models' 0\% sequence accuracies. (m)BART struggles more with syntactic transformations as indicated by its lower average accuracies, though it is still capable of detecting the correct auxiliaries and objects to move as indicated by the high \emph{maximum} main auxiliary and object noun accuracies in Figure~\ref{fig:all_models_scatter}. This indicates that \textbf{pre-trained seq2seq models demonstrate a hierarchical inductive bias}, and that they can quickly learn syntactic transformations.

\begin{table}[]
    \centering
    \resizebox{\linewidth}{!}{
    \begin{tabular}{lrrrr}
    \toprule
     & \multicolumn{2}{c}{\bf{Question Formation}} & \multicolumn{2}{c}{\bf{Passivization}} \\\cmidrule(lr){2-3}\cmidrule(lr){4-5}
    \bf{Model} & English & German & English & German \\
    \midrule
    T5 & 0.48 & -- & 0.25 & -- \\
    mT5 & 0.50 & 0.44 & 0.25 & 0.50 \\
    BART & 0.40 & -- & 0.30 & -- \\
    mBART & 0.48 & 0.38 & 0.29 & 0.44 \\
    \bottomrule
    \end{tabular}}
    \caption{\emph{Maximum} main auxiliary and object noun accuracies through 500 epochs of fine-tuning \emph{after randomizing the weights} of each pre-trained model. Sequence accuracies remain near 0 throughout fine-tuning.}
    \label{tab:randomized}
\end{table}

There are two main differences between the two classes of models we test: (m)T5 and (m)BART are not only pre-trained, but are also much deeper and much more parameterized than our non-pre-trained models. Are hierarchical inductive biases a feature of deep architectures, then, or are they acquired during pre-training? To control for pre-training while keeping the model size consistent, we randomize the weights of mT5 (the better-performing model) and fine-tune for up to 500 epochs using an initial LR\footnote{We tune over learning rates $\in 5\times 10^{\{-2,-3,-4,-5\}}$ for the randomized models, finding that $5\times 10^{-4}$ yields the best main auxiliary and object noun accuracies on in-domain evaluations.} of $5\times 10^{-4}$. For all of the transformations, the maximum accuracies of the randomized models are much lower than the average accuracies of the pre-trained models (Table~\ref{tab:randomized}), which suggests that the deeper architecture on its own does not lead to structure-sensitive generalizations.  This in return indicates that \textbf{pre-trained models do not start with a hierarchical inductive bias; they \emph{acquire} it through pre-training}, extending the findings of \citet{warstadt2020linguistic} to generative sequence-to-sequence models. 
However, as indicated by the non-zero main auxiliary/object noun accuracies, the randomly initialized mT5 models do not exhibit a consistent linear generalization either---unlike the 1-layer non-pre-trained models. This may be due to the large number of parameters compared to the size of the transformations training corpus. A randomly initialized model of this size would likely need orders of magnitude more training data to learn stable generalizations.

Each pre-trained model almost always chooses the correct auxiliary/object to move; what errors account for their sub-perfect sequence accuracies, then? We perform a detailed error analysis, finding that pre-trained models drop PPs from the second noun phrase but otherwise perform many complex hierarchy-sensitive transformations properly. See Appendix~\ref{app:error_analysis} for details.

\section{Transformation Strategies}\label{sec:strategies}
Our results indicate that pre-trained seq2seq models can consistently perform hierarchy-sensitive transformations. What strategy do they follow to do this? Because pre-training corpora include actives, passives, declaratives, and questions, model representations could encode these high-level sentence features.\footnote{For example, (sets of) neuron activations have been found to encode syntactic features in MLMs \citep{ravfogel2021counterfactual,finlayson2021causal,hernandez2021dimensional}.}  Thus, one strategy could be to learn a mapping between abstract representations of different sentence structures  (\textsc{representation} strategy). Alternatively, models could learn to correctly identify the relevant syntactic units in the input, and then learn a ``recipe'' of steps leading to the correct transformations (\textsc{recipe} strategy).

To distinguish which strategy models use to perform syntactic transformations, we observe cross-lingual zero-shot transfer on syntactic transformations. We exploit that English and German use the same operations for question formation, whereas passivization in German involves the additional steps of case reinflection and moving the main verb. If structural representations are shared across English and German,\footnote{Shared cross-lingual structural representations have been found for multilingual MLMs \cite{chi-etal-2020-finding}, and we provide further evidence for shared representations in this section.} we do not expect divergent behaviors for question formation and passivization: if a model employs the \textsc{representation} strategy, then after fine-tuning on only English passivization, it should also correctly perform German passivization, including the additional steps of case reinflection and moving the main verb. Conversely, if it employs the \textsc{recipe} strategy, we expect a model trained on English passivization to only perform the steps that are required for English passivization, resulting in incorrect case marking and no main verb movement in German.

\begin{figure}
    \centering
    \includegraphics[width=0.49\linewidth]{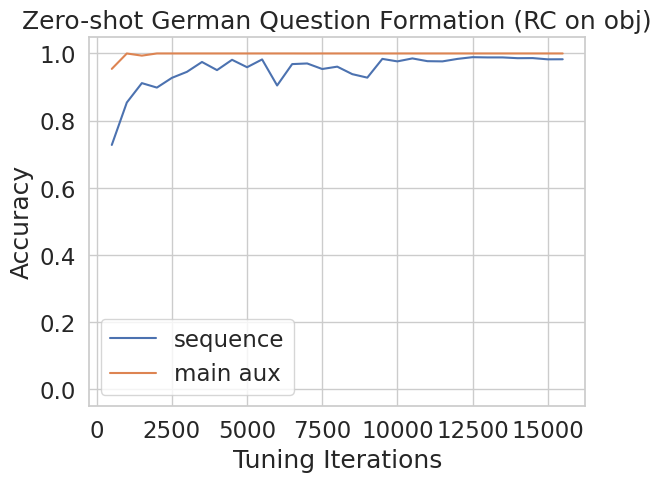}
    \hfill
    \includegraphics[width=0.49\linewidth]{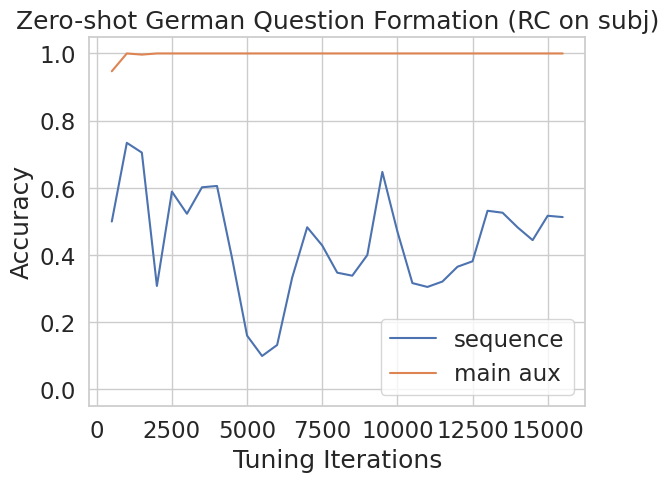}
    \newline
    \includegraphics[width=0.49\linewidth]{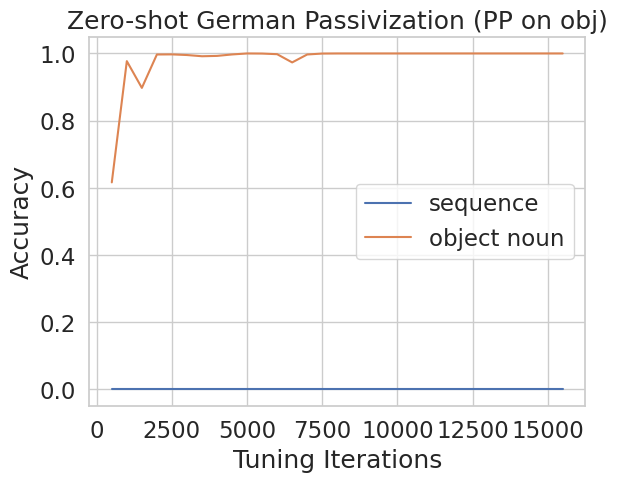}
    \hfill
    \includegraphics[width=0.49\linewidth]{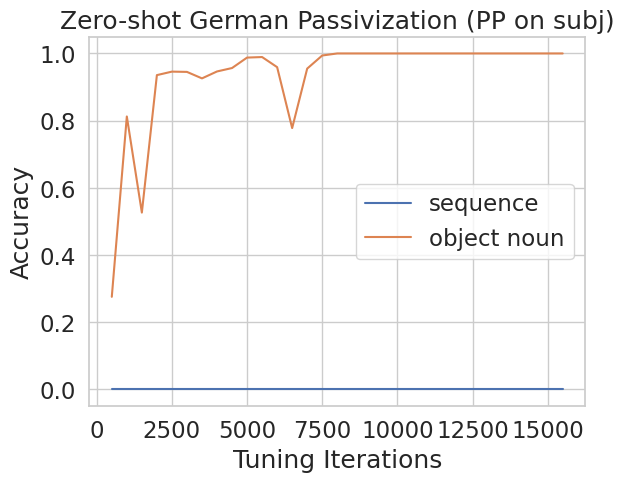}
    \caption{Learning curves for mT5 on German transformations after fine-tuning on English/German identity examples and English transformations. We show accuracies for German question formation with RCs on objects (top left) and RCs on subjects (top right), as well as accuracies for German passivization with PPs on objects (bottom left) and PPs on subjects (bottom right).}
    \label{fig:zeroshot_question_de}
\end{figure}

We first verify that mT5 and mBART are capable of cross-lingual transfer by training a model on the English question formation task and evaluating on German. In early experiments, we noticed the issue of ``spontaneous translation'' \cite{xue2021mt5}; we therefore also include German identity transformations in the training data to train the decoder to also output German sentences. 

As the top two panels of Figure~\ref{fig:zeroshot_question_de} show, mT5 can correctly perform German question formation on in-domain structures (RCs on objects) after being exposed only to English transformations. For out-of-domain structures (RCs on subjects), mT5 almost always moves the main auxiliary but almost never deletes it from its original position, resulting in lower sequence accuracies. Apart from this error, the model is capable of cross-lingual transfer on the question formation task. By contrast, mBART achieves poor results on zero-shot German question formation, so we cannot make conclusive arguments using this approach; see Appendix~\ref{app:zeroshot_mbart}.

Given that cross-lingual transfer is possible for mT5, how does the model behave in the passivization task, which differs between English and German? We fine-tune mT5 on English passivization (as well as German identity transformations on active sentences). The results of this experiment (the lower two panels in Figure~\ref{fig:zeroshot_question_de}) show that the model is still able to move the main object to the subject position, but also that it never correctly performs German passivization in its entirety. This is because the model performs exactly the same steps for German sentences as for English sentences, resulting in outputs with English syntax:

\exg. \textcolor{red}{Meinen} Kater bei ihrem Molch \textcolor{red}{was} \textcolor{red}{verwirrten} \textcolor{red}{by} \textcolor{red}{die} Esel.\\
    My.\textsc{\textcolor{red}{acc}} cat by your.\textsc{dat} newt \textcolor{red}{was} confused.\textsc{\textcolor{red}{past}} \textcolor{red}{by} the.\textsc{\textcolor{red}{nom}} donkeys.\\

These behavioral patterns suggest that mT5 employs the \textsc{recipe} strategy: it succeeds if a transformation's required operations are the same across languages (as for question formation) but fails if the steps differ (as for passivization). Even in passivization, however, the model still learns to move the correct NPs, which provides additional evidence that mT5 makes use of structural features when performing transformations. Given the similarities between mT5's and mBART's architectures and training setups, one could reasonably presume that mBART may follow a similar strategy to perform syntactic transformations; nonetheless, mBART is less consistent in performing syntactic transformations, so this method cannot present strong evidence for use of the \textsc{recipe} strategy for that model.

\section{Corpus Analysis}
Pre-trained models learn to use hierarchical features for performing syntactic transformations. Is this because there is explicit supervision for the hierarchical rules in the pre-training corpora? In other words, are there \emph{disambiguating examples} in these models' training corpora that helps them memorize hierarchical transformation patterns? Here, we focus on English question formation examples in mT5's training corpus.\footnote{mT5 outperforms mBART. If disambiguating contexts in the pre-training data lead to syntactic generalizations, then we expect these examples to be more likely in mT5's training corpus.} Disambiguating examples would be rare, as a single pre-training context window must contain a declarative sentence as well as the same sentence transformed into a question; humans would tend to replace at least some of the constituents with pronouns or delete them \cite[e.g.,][]{ariel2001accessibility}. It would also require the \textsc{move-first} rule to not correctly transform the sentence---\emph{and} for at least one of the auxiliaries to be noised in one sentence but not the other, such that the auxiliary has to be recovered from the other sentence. For example:

\ex.\label{ex:disambiguating}\ldots\textcolor{blue}{Has} this company which \textcolor{red}{hasn't} had any legal violations been reported to the Better Business Bureau? This company which \textcolor{red}{hasn't} had any legal violations \textcolor{blue}{\texttt{<X>}} been reported to the Better Business Bureau\ldots

We search for English disambiguating question formation examples. To this end, we sample $5$M English documents from mT5's training corpus mC4, segmenting each document into sentences using spaCy.\footnote{\url{https://spacy.io}} This yields $118.3$M sentences. We examine each pair of adjacent sentences in each document, manually inspecting any sentence pair meeting the following criteria: (1) the token Jaccard similarity of the sentences is $>0.7$; (2) one sentence begins with an auxiliary verb and the other does not; (3) there are at least two distinct auxiliaries in both sentences. There are $277$ sentence pairs in our sample that met all criteria, of which $13$ are adjacent declarative/question pairs that are equivalent except for the fronted auxiliary. Thus, the probability of an equivalent declarative/question pair with two auxiliaries in mC4 is $\approx 1.1 \times 10^{-7}$. As T5's and mBART's training corpora consist of data from similar webtext distributions, it is likely that these structures exist in those corpora as well.

Crucially, however, none of the declarative/question pairs were disambiguating examples: each pair was consistent with the linear \textsc{move-first} rule. What is the probability of a disambiguating example, then? If we assume that the probability of a sentence containing an RC on the subject is independent from the probability of a declarative/question sentence pair, we can take the product of both probabilities to obtain an estimate. From the same sample of $118.3$M sentences, we use spaCy's dependency parser to extract sentences containing an RC on the subject and where at least one auxiliary verb appears in the sentence. We obtain $526,944$ such sentences, meaning that the probability of an RC on a subject in an auxiliary-containing sentence in mC4's English corpus is $\approx 4.5\times 10^{-3}$. Thus, the probability of declarative/question pair with an RC on the subject and auxiliary in the RC is $\approx (4.5\times 10^{-3}) \cdot (1.1\times 10^{-7}) = 4.95\times 10^{-10}$. mT5 is trained on up to 1T tokens of data, and $5.67$\% of its documents are English; it therefore observes $\approx 56.7$B English tokens. If we optimistically assume that English sentences contain an average of $15$ tokens, it observes $3.78$B English sentences. Then we would expect $3.78$B$ \times (4.95\times 10^{-10}) \approx 2$ disambiguating examples. This is not including the auxiliary masking criterion, which would make such examples even less likely.

Thus, while we cannot definitively rule out the possibility of disambiguating examples in mC4, they are rare if they exist in the corpus at all. Nonetheless, we have found evidence for supervision on question formation in the form of adjacent declarative/question sentence pairs, even if they do not explicitly support the hierarchical rule.

\section{Discussion}
Our experiments provide evidence that pre-trained seq2seq models acquire a hierarchical inductive bias through exposure to non-annotated natural language text. This extends the findings of \citet{warstadt2020linguistic} and \citet{warstadt2020learning} to a more challenging generative task, where models cannot rely on n-gram distributional heuristics \citep{kam2008bigrams}. This also provides additional evidence that masking and reconstructing subsets of input sequences is a powerful training objective for inducing linguistic generalizations, whether in masked language models like RoBERTa \citep{warstadt2020linguistic} or sequence-to-sequence models. Span denoising ((m)T5's objective) appears more effective for learning syntactic transformations than full sequence reconstruction ((m)BART's objective) given that (m)T5 is more consistently able to perform transformations, though there are too many other differences in training data and hyperparameters between (m)T5 and (m)BART for us to be able to directly implicate the training objective. This hypothesis can be tested explicitly in future work by training identical models that differ only in their pre-training objective. 

Counter to \citet{mccoy2020trees}, our findings suggest that hierarchical architectural constraints (e.g., tree-structured networks) are not necessary for robust hierarchical generalization as long as the model has been exposed to large amounts of natural language text---possibly far more language than humans would be exposed to. However, one difference between the randomly initialized models employed by \citet{mccoy2020trees} and pre-trained models is that pre-trained models have likely seen the structures (but not sentences) present in the generalization set; thus, rather than relying on syntactic features, the model could choose the correct transformation because it is more similar to the grammatical examples it has already seen. We found declarative/question pairs in mT5's training corpus, but we did not find any examples that explicitly demonstrated the hierarchical rule for question formation. While we cannot fully rule out the possibility of disambiguating examples, this strategy is still unlikely given that pre-trained models produce ungrammatical transformations, both in monolingual transformations (e.g., not deleting the main auxiliary after copying it to the start of the sentence) and in cross-lingual German passivization. Additionally, because we use greedy decoding, models are not able to take future words into account when predicting the fronted auxiliary: they must select the appropriate auxiliary to move solely based on the encoder's representations.

More broadly, our findings counter the assumption that a hierarchical constraint is necessary in language learners to acquire hierarchical generalization \citep{chomsky1965aspects}. While the pre-trained models that we considered observe far more input than a child would receive \citep{linzen2020accelerate}, \citet{huebner2021babyberta} recently demonstrated high performance on grammaticality judgments for models trained on much smaller child-directed speech corpora, suggesting that our findings may also hold when training models on more human-like input.

\section{Conclusions}
We have performed an analysis of the syntactic transformational ability of large pre-trained sequence-to-sequence models. We find that pre-trained models acquire a hierarchical inductive bias during pre-training, and that the architecture does not yield this hierarchical bias by itself.

It remains an open question whether such deep and highly parameterized models or such large pre-training datasets are necessary for hierarchical generalization. Future work could ablate over model depth and pre-training corpus size to observe the relative contribution of architecture and the training set to inducing hierarchical inductive biases in seq2seq models.

\section*{Acknowledgments}
We thank the members of NYU's Computation and Psycholinguistics Lab and the reviewers for their thoughtful feedback. We also thank R. Thomas McCoy for providing sentence generation scripts.

This material is based upon work supported by the National Science Foundation (NSF) under
Grant \#BCS-2114505, \#BCS-1919321, as well as Grant \#2030859 to the Computing Research Association for the CIFellows Project. Aaron Mueller was supported by an NSF Graduate Research Fellowship (Grant  \#1746891).

\bibliography{anthology,custom}
\bibliographystyle{acl_natbib}

\clearpage

\appendix

\section{German Structures}\label{app:train_test_gen_de}
Here, we present examples of the sentences in the training, development, test, and generalization sets for the German question formation and passivization tasks (Table~\ref{tab:train_test_gen_de}). As in English, we train the model on declarative or active sentences, as well as question-formation or passivization examples with no RCs/PPs or with RCs/PPs on subjects (i.e., sentences that are consistent with the hierarchical and linear rules described in \S\ref{sec:data}). Then we evaluate its generalization on sentences where the linear rule does not properly transform the sentence.

For further clarity, we present glossed examples of each German structure below for both tasks.

\ex.\textit{German Question Formation (no RC)}:\label{ex:qform_de_no_rc}
    \ag. Unsere Salamander haben die Pfaue bewundert.\\
        Our.\textsc{nom} salamanders have the.\textsc{acc} peacocks admired.\\
        "Our salamanders have admired the peacocks."
    \bg. Haben unsere Salamander die Pfaue bewundert?\\
        Have our.\textsc{nom} salamanders the.\textsc{acc} peacocks admired?\\
        "Have our salamanders admired the peacocks?"
        
\ex.\textit{German Question Formation (RC on object)}:\label{ex:qform_de_rc_obj}
    \ag. Einige Molche können meinen Papagei, der deinen Raben trösten kann, nerven. \\
        Some.\textsc{nom} newts can my.\textsc{acc} parrot, that.\textsc{nom} your.\textsc{acc} ravens comfort can, annoy.\\
        "Some newts can annoy my parrot that can comfort your ravens."
    \bg. Können einige Molche meinen Papagei, der deinen Raben trösten kann, nerven? \\
        Can some.\textsc{nom} newts my.\textsc{acc} parrot, that.\textsc{nom} your.\textsc{acc} ravens comfort can, annoy?\\
        "Can some newts annoy my parrot that can comfort your ravens?"

\ex.\textit{German Question Formation (RC on subject)}:\label{ex:qform_de_rc_subj}
    \ag. Ihr Hund, den ihr Geier nerven kann, hat einige Pfauen amüsiert. \\
        Your.\textsc{nom} dog, that.\textsc{acc} your.\textsc{nom} vulture annoy can, has some.\textsc{acc} peacocks amused.\\
        "Your dog that can annoy your vulture has amused some peacocks."
    \bg. Hat ihr Hund, den ihr Geier nerven kann, hat einige Pfauen amüsiert. \\
        Has your.\textsc{nom} dog, that.\textsc{acc} your.\textsc{nom} vulture annoy can, some.\textsc{acc} peacocks amused?\\
        "Has your dog that can annoy your vulture amused some peacocks?"

\ex.\ \textit{German Passivization (no PP)}:\label{ex:passiv_de_no_pp}
    \ag. Ihr Kater bedauerte den Dinosaurier. \\
        Your.\textsc{nom} cat pities the.\textsc{acc} dinosaur.\\
        "Your cat pities the dinosaur."
    \bg. Der Dinosaurier wurde von ihrem Kater bedauert. \\
        The.\textsc{nom} dinosaur was from your.\textsc{dat} cat pitied.\\
        "The dinosaur was pitied by your cat."
        
\ex.\ \textit{German Passivization (PP on object)}:\label{ex:passiv_de_pp_obj}
    \ag. Unsere Ziesel amüsierten einen Kater hinter dem Dinosaurier. \\
        Our.\textsc{nom} ground-squirrels amuse a.\textsc{acc} cat behind the.\textsc{dat} dinosaur.\\
        "Our ground squirrels amuse a cat behind the dinosaur."
    \bg. Ein Kater hinter dem Dinosaurier wurde von unseren Zieseln amüsiert. \\
        A.\textsc{nom} cat behind the.\textsc{dat} dinosaur was from our.\textsc{dat} ground-squirrels amused.\\
        "A cat behind the dinosaur was amused by our ground squirrels."
        
\ex.\ \textit{German Passivization (PP on subject)}:\label{ex:passiv_de_pp_subj}
    \ag. Die Geier hinter meinem Ziesel akzeptieren die Molche.\\
        The.\textsc{nom} vultures behind my.\textsc{dat} ground-squirrel accept the.\textsc{acc} newts.\\
        "The vultures behind my ground squirrel accept the newts."
    \bg. Die Molche wurden von den Geiern hinter meinem Ziesel akzeptiert.\\
        The.\textsc{nom} newts were from the.\textsc{dat} vultures behind my.\textsc{dat} ground-squirrel accepted.\\
        "The newts were accepted by the vultures behind my ground squirrel."

\begin{table*}[]
    \centering
    \resizebox{0.8\linewidth}{!}{
    \begin{tabular}{lp{6cm}p{8cm}}
    & \fcolorbox{black}{white}{\phantom{\vrule width 1.25mm height 1.25mm}} Train, dev, test & \fcolorbox{black}{cellgray}{\phantom{\vrule width 1.25mm height 1.25mm}} Generalization \\
    \toprule
    \textbf{Question Formation} & Declarative & Question \\\midrule
    No RC & decl: unsere Salamander haben die Pfaue bewundert. \newline$\rightarrow$ unsere Salamander haben die Pfaue bewundert. & quest: ihre Hunde haben unseren Orang-Utan genervt. \newline$\rightarrow$ haben ihre Hunde unseren Orang-Utan genervt? \\\midrule
    RC on object & decl: unser Ziesel kann den Salamander, der meinen Pfau verwirrt hat, akzeptieren. \newline$\rightarrow$ unser Ziesel kann den Salamander, der meinen Pfau verwirrt hat, akzeptieren. & quest: einige Molche können meinen Papagei, der deinen Raben trösten kann, nerven. \newline$\rightarrow$ können einige Molche meinen Papagei, der deinen Raben trösten kann, nerven? \\\midrule
    RC on subject & decl: dein Molch, den mein Wellensittich bewundert hat, kann meine Dinosaurier trösten. \newline$\rightarrow$ dein Molch, den mein Wellensittich bewundert hat, kann meine Dinosaurier trösten. & \cellcolor{cellgray} quest: ihr Hund, den ihr Geier nerven kann, hat einige Pfaue amüsiert. \newline$\rightarrow$ hat ihr Hund, den ihr Geier nerven kann, einige Pfaue amüsiert? \\
    \midrule
    \midrule
    \textbf{Passivization} & Active & Passive \\\midrule
    No PP & decl: die Löwen unterhielten einen Wellensittich. \newline$\rightarrow$ die Löwen unterhielten einen Wellensittich. & passiv: ihr Kater bedauerte den Dinosaurier. \newline$\rightarrow$ der Dinosaurier wurde von ihrem Kater bedauert. \\\midrule
    PP on object & decl: ihre Geier verwirrten ihren Raben über unserem Ziesel. \newline$\rightarrow$ ihre Geier verwirrten ihren Raben über unserem Ziesel. & passiv: unsere Ziesel amüsierten einen Kater hinter dem Dinosaurier. \newline$\rightarrow$ ein Kater hinter dem Dinosaurier wurde von unseren Zieseln amüsiert. \\\midrule
    PP on subject & decl: ein Löwe unter unserem Hund nervte einigie Ziesel. \newline$\rightarrow$ ein Löwe unter unserem Hund nervte einigie Ziesel. & \cellcolor{cellgray} passiv: die Geier hinter meinem Ziesel akzeptieren die Molche. \newline$\rightarrow$ die Molche wurden von den Geiern hinter meinem Ziesel akzeptiert. \\
    \bottomrule
    \end{tabular}}
    \caption{The distribution of syntactic structures in the German train, test, and generalization sets. We use the test set to evaluate whether models have learned the task on in-distribution examples, and the generalization set to evaluate hierarchical generalization.}
    \label{tab:train_test_gen_de}
\end{table*}

\section{Learning Curves}\label{app:learning_curves}
\begin{figure*}
    \centering
    \includegraphics[height=3.5cm]{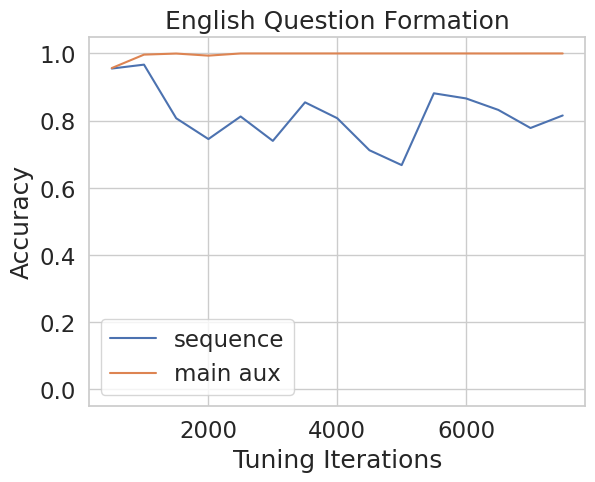}
    \hfill
    \includegraphics[height=3.5cm,trim={2cm 0 0 0},clip]{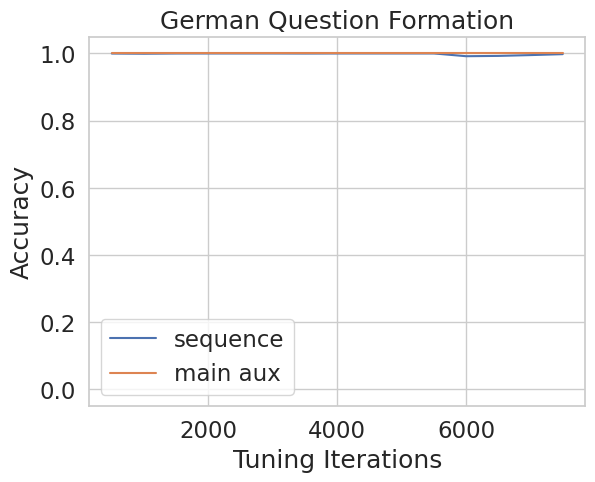}
    \hfill
    \includegraphics[height=3.5cm,trim={2cm 0 0 0},clip]{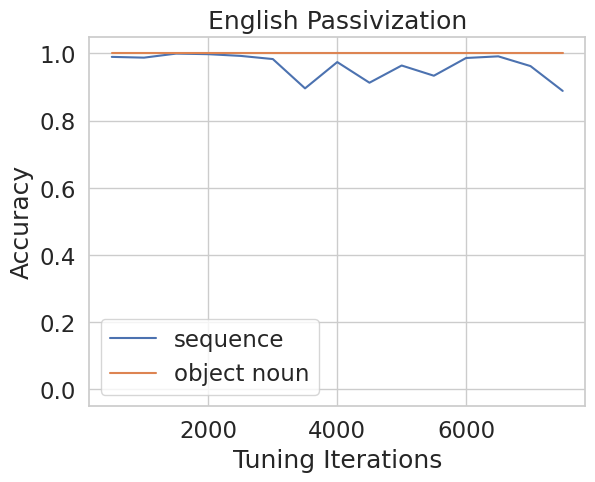}
    \hfill
    \includegraphics[height=3.5cm,trim={2cm 0 0 0},clip]{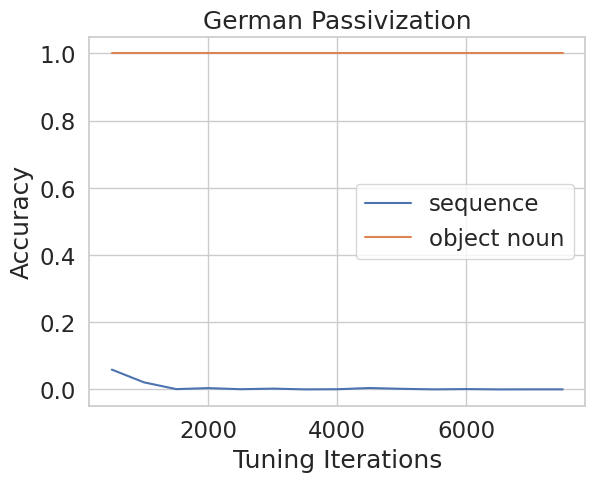}
    \caption{Learning curves over 10 epochs of fine-tuning for mT5 on both syntactic transformation tasks.}.
    \label{fig:mt5_curves}
\end{figure*}

\begin{figure*}
    \centering
    \includegraphics[height=3.5cm]{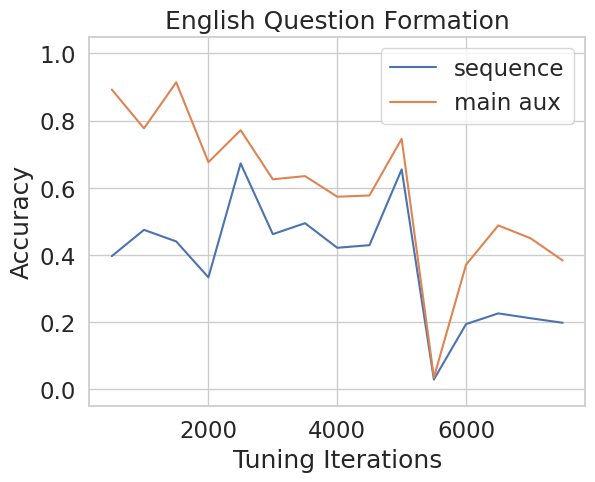}
    \hfill
    \includegraphics[height=3.5cm,trim={2cm 0 0 0},clip]{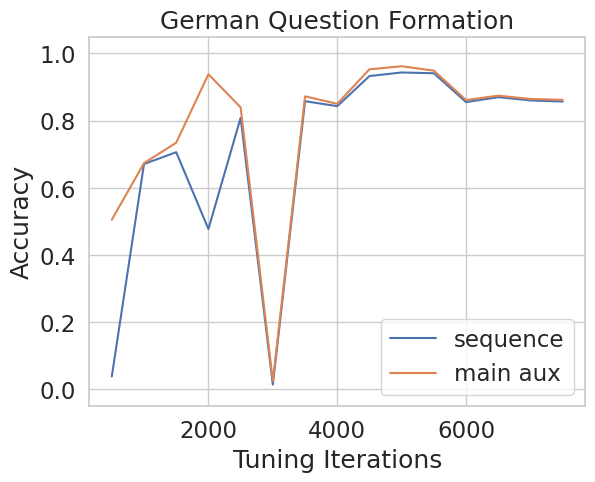}
    \hfill
    \includegraphics[height=3.5cm,trim={2cm 0 0 0},clip]{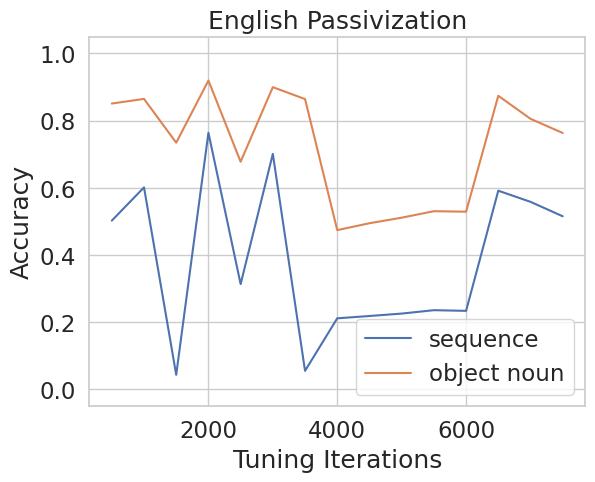}
    \hfill
    \includegraphics[height=3.5cm,trim={2cm 0 0 0},clip]{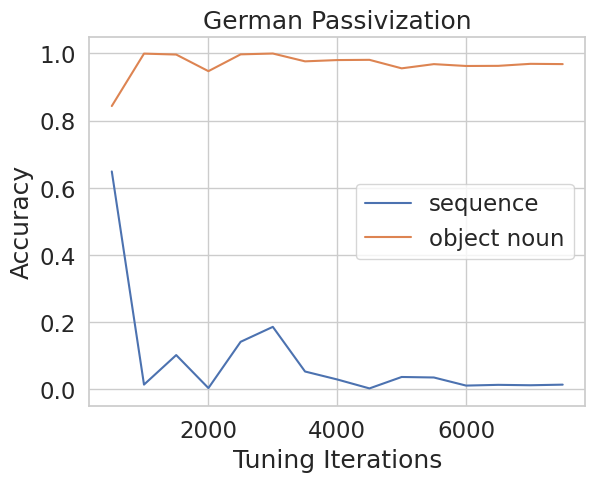}
    \caption{Learning curves over 10 epochs of fine-tuning for mBART on both syntactic transformation tasks.}.
    \label{fig:mbart_curves}
\end{figure*}

\begin{figure}
    \centering
    \includegraphics[height=3.2cm]{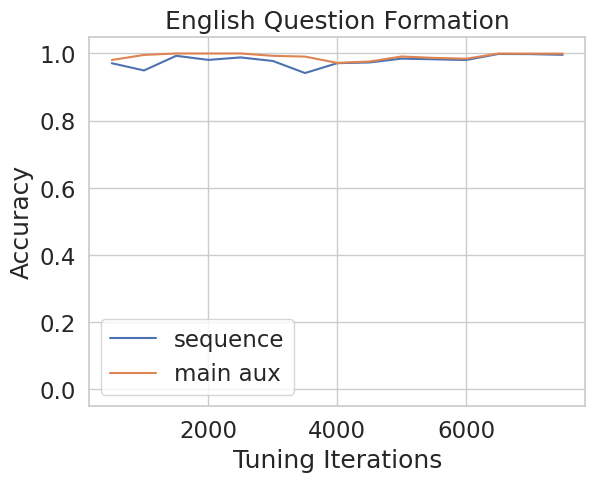}
    \hfill
    \includegraphics[height=3.2cm,trim={2cm 0 0 0},clip]{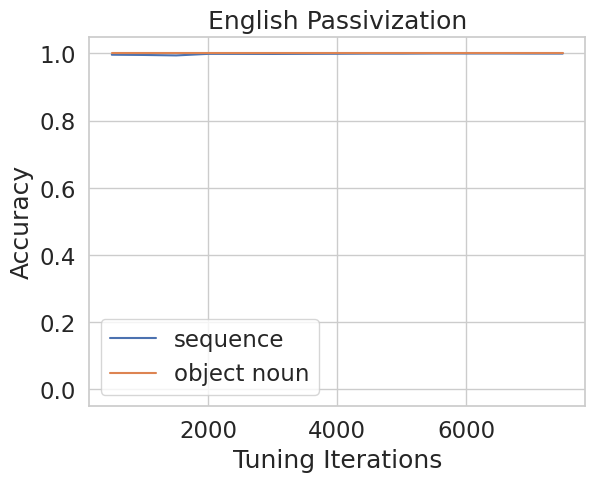}
    \caption{Learning curves over 10 epochs of fine-tuning for T5 on both syntactic transformation tasks.}
    \label{fig:t5_curves}
\end{figure}

\begin{figure}
    \centering
    \includegraphics[height=3.2cm]{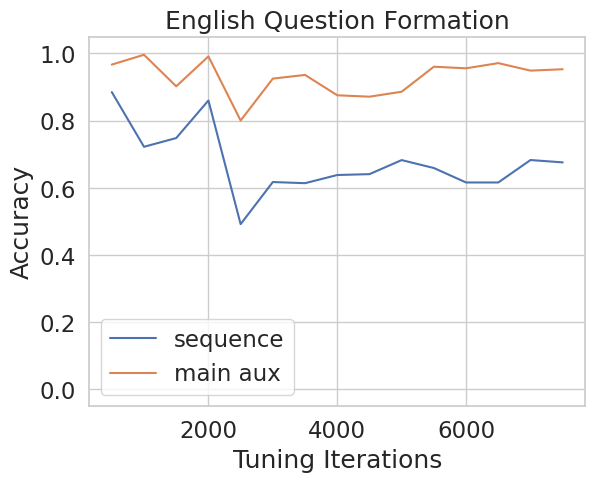}
    \hfill
    \includegraphics[height=3.2cm,trim={2cm 0 0 0},clip]{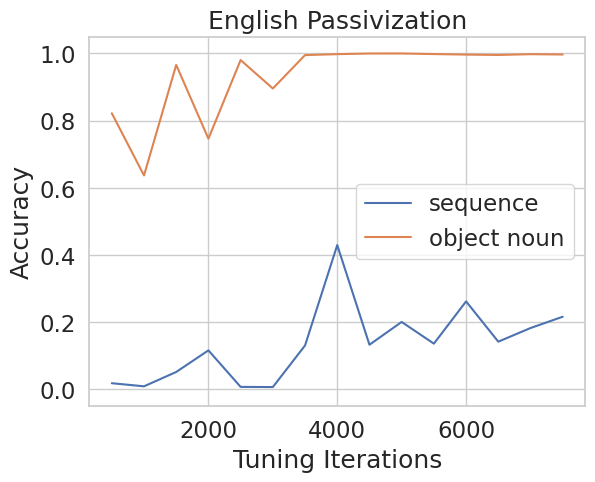}
    \caption{Learning curves over 10 epochs of fine-tuning for BART on both syntactic transformation tasks.}
    \label{fig:bart_curves}
\end{figure}

Here, we present learning curves for 10 epochs of fine-tuning for each transformation in each language (Figures~\ref{fig:mt5_curves},\ref{fig:mbart_curves},\ref{fig:t5_curves},\ref{fig:bart_curves}). The accuracies shown in these curves are the same as those shown in Figure~\ref{fig:all_models_scatter}, but now associated with their respective fine-tuning iteration.

All models except mBART immediately achieve near-perfect main auxiliary and object noun accuracies. Their loss on the validation sets converges almost immediately, so it's possible that reductions in generalization accuracies throughout fine-tuning are due to overfitting to the training distribution. For mBART, however, main auxiliary and object noun accuracies start high and then begin to vary dramatically throughout fine-tuning. This is perhaps due to quick overfitting on the training distribution. We analyze what errors cause mBART's deficiencies in \S\ref{app:mbart_error_analysis}.

\section{Error Analysis}\label{app:error_analysis}
Each pre-trained model almost always chooses the correct auxiliary/object to move; what other errors account for their sub-perfect sequence accuracies? We implement more specific metrics to observe more closely what mistakes (m)T5 and (m)BART are making. We show results for mT5 in \S\ref{app:mt5_error_analysis} and mBART in \S\ref{app:mbart_error_analysis}, but the errors we discuss are generally consistent across models. We also present more detailed metrics for the most complex transformation, German passivization, in \S\ref{app:more_detailed_error_analysis}.

\subsection{mT5}\label{app:mt5_error_analysis}

\begin{figure}
    \centering
    \includegraphics[width=0.49\linewidth]{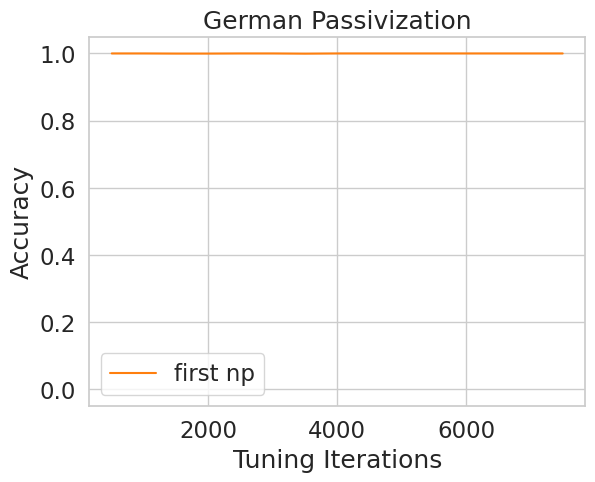}
    \hfill
    \includegraphics[width=0.49\linewidth]{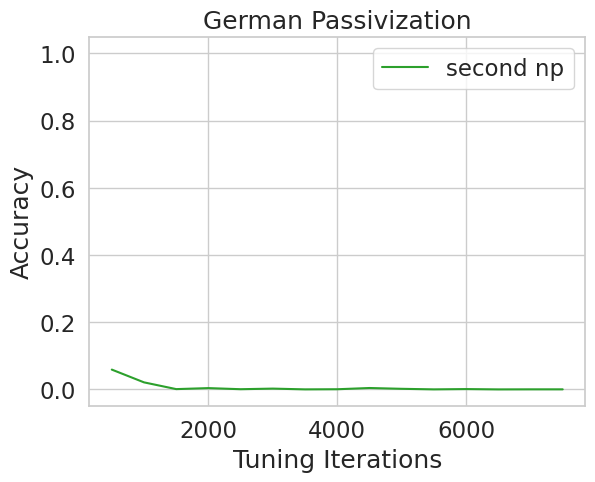}
    \newline
    \includegraphics[width=0.49\linewidth]{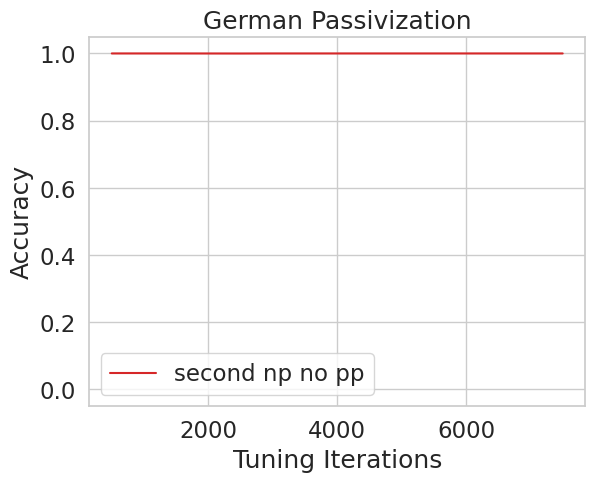}
    \hfill
    \includegraphics[width=0.49\linewidth]{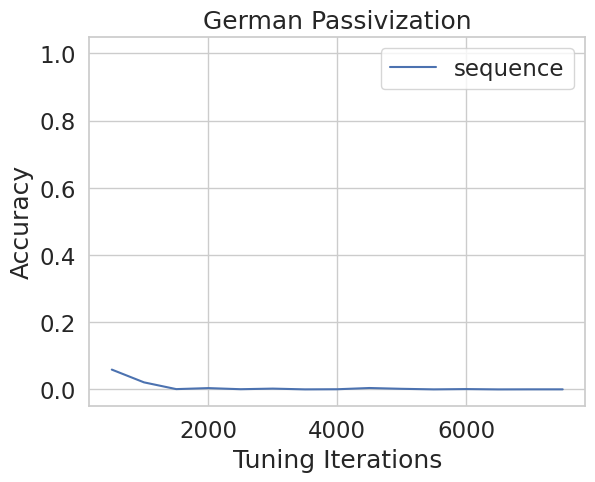}
    \caption{Learning curves displaying alternative accuracy metrics for mT5 on German passivization. We present the accuracy of the model in properly moving the object NP to the start of the sentence (top left), moving the subject NP after the auxiliary verb (top right), moving the subject NP after the auxiliary verb with or without its attached PP (bottom left), and the full sequence accuracy (bottom right).}
    \label{fig:mt5_error_analysis}
\end{figure}

Figure~\ref{fig:mt5_error_analysis} depicts results for mT5 for German passivization, the transformation on which all models achieve the lowest sequence accuracy. mT5 is almost always successful at the hierarchical transformation of moving the object NP to subject position (including its attached PP when present), and it correctly moves the original subject noun to a ``by'' phrase following the auxiliary. However, for both English and German passivization, the main error accounting for sub-perfect sequence accuracies is that the model fails to
preserve the PP on the second NP (in the by-phrase):

\ex.\ \ My yaks \textcolor{orange}{below the unicorns} comforted the orangutans.\newline
    $\rightarrow$ The orangutans were comforted by my yaks.

As mT5 has not been fine-tuned on output sequences where PPs appear at the end of the sentence, the decoder could be assigning low probabilities to end-of-sentence PPs while otherwise encoding a hierarchical analysis of sentence structure.

Errors for question formation are more varied. Pre-trained models' sub-perfect main auxiliary accuracies on question formation are mainly due to improper negations on the main auxiliary: when the noun in the relative clause and the main noun agree in number, models will sometimes delete the main auxiliary (as expected) while copying the incorrect auxiliary to the beginning of the sentence. Additionally, the discrepancy between sequence and main auxiliary accuracies is almost always attributable to models not deleting the main auxiliary after moving it to the start of the sentence. These results (as with the passivization results) suggest that \textbf{pre-trained seq2seq models are better at performing hierarchy-sensitive transformations than the sequence accuracies initially suggest---but also that they can fail to perform theoretically simpler operations,} such as deletions and moving all parts of a constituent.

We present more detailed error analyses in Appendix~\ref{app:more_detailed_error_analysis}. We find that pre-trained models also consistently succeed in case reinflection, tense reinflection, and passive auxiliary insertion.

\subsection{mBART}\label{app:mbart_error_analysis}
\begin{figure}
    \centering
    \includegraphics[width=0.49\linewidth]{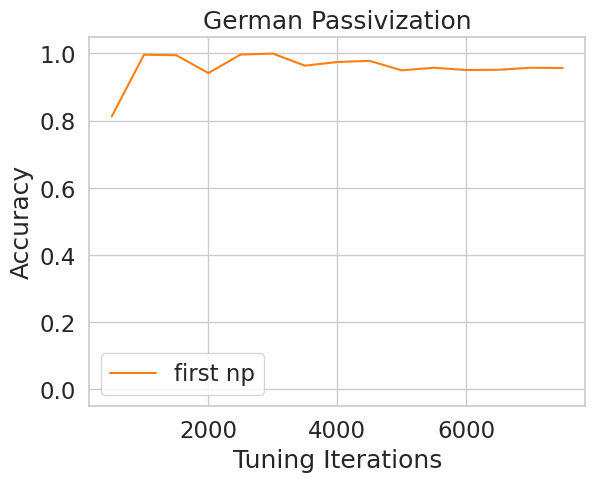}
    \hfill
    \includegraphics[width=0.49\linewidth]{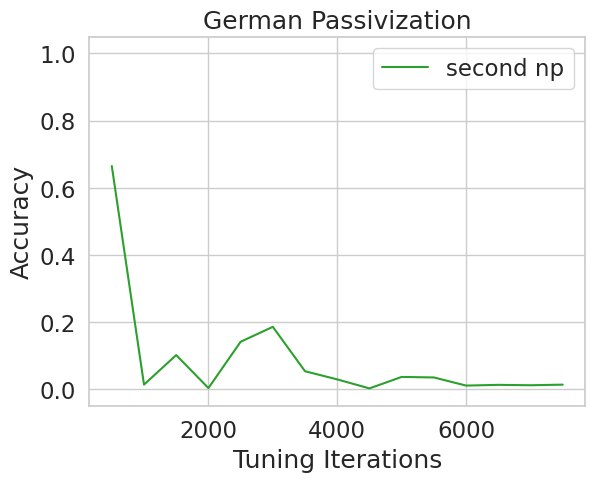}
    \newline
    \includegraphics[width=0.49\linewidth]{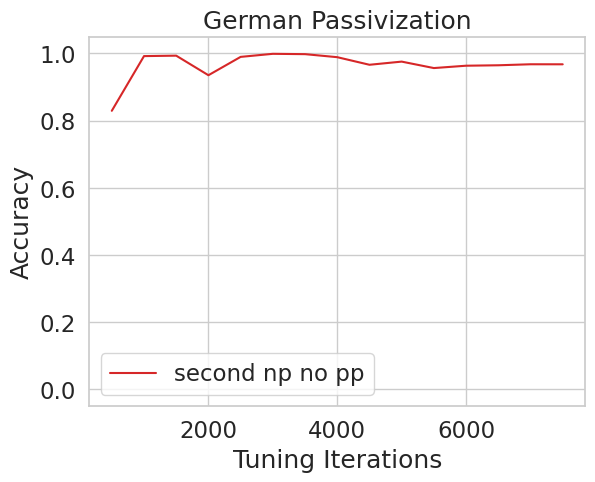}
    \hfill
    \includegraphics[width=0.49\linewidth]{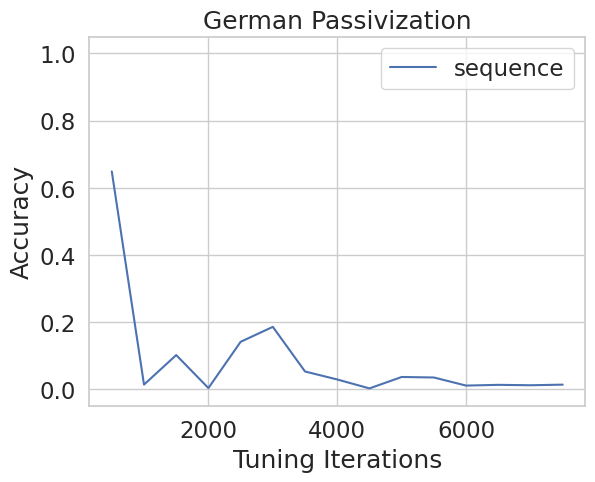}
    \caption{Learning curves displaying alternative accuracy metrics for mBART on German passivization. We present the accuracy of the model in properly moving the object NP to the start of the sentence (top left), moving the subject NP after the auxiliary verb (top right), moving the subject NP after the auxiliary verb with or without its attached PP (bottom left), and the full sequence accuracy (bottom right).}
    \label{fig:mbart_error_analysis}
\end{figure}

We have shown in \S\ref{app:mt5_error_analysis} that mT5 achieves sub-perfect sequence accuracies on passivization due to its dropping the prepositional phrase on the second NP. Here, we present results for mBART (Figure~\ref{fig:mbart_error_analysis}). The takeaways for mBART are similar to mT5's: the model succeeds in moving the proper nouns, but it often drops the prepositional phrase from the second NP during movement.

As the model also fails to perform English question formation consistently, we also observe what errors it makes in that task. We find that deficiencies in main auxiliary accuracy are due to the model copying the incorrect auxiliary to the beginning of the sentence, while gaps between main auxiliary and sequence accuracy are due to the model dropping the relative clause on the second NP.

\subsection{More Detailed Metrics}\label{app:more_detailed_error_analysis}
\begin{figure}
    \centering
    \includegraphics[width=0.49\linewidth]{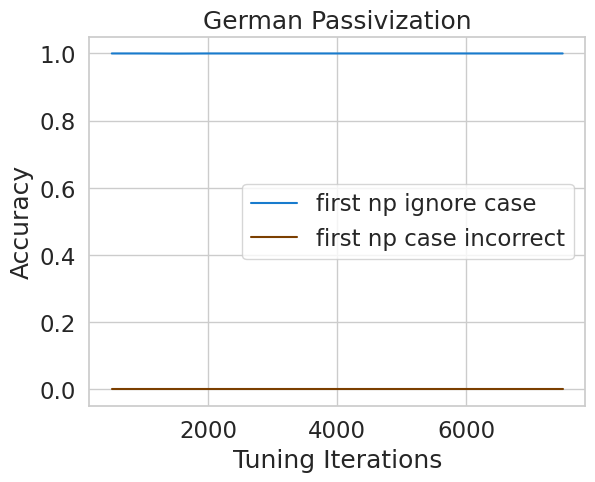}
    \hfill
    \includegraphics[width=0.49\linewidth]{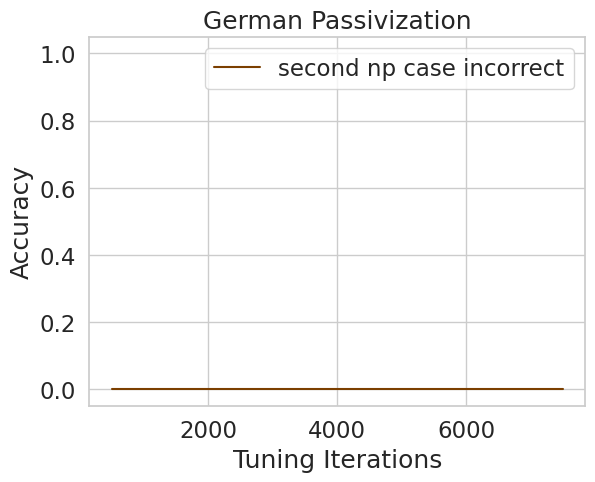}
    \newline
    \includegraphics[width=0.49\linewidth]{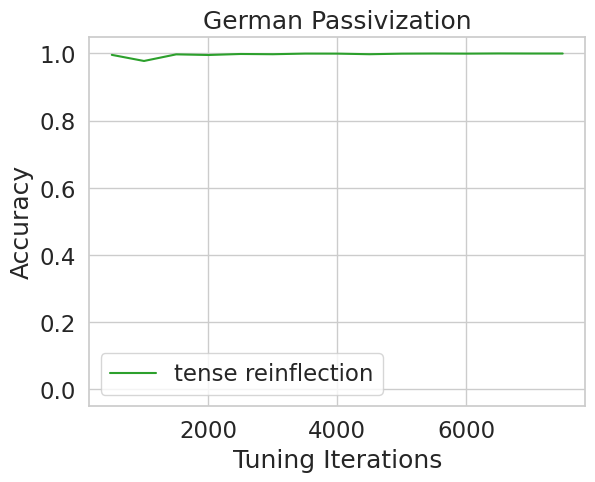}
    \hfill
    \includegraphics[width=0.49\linewidth]{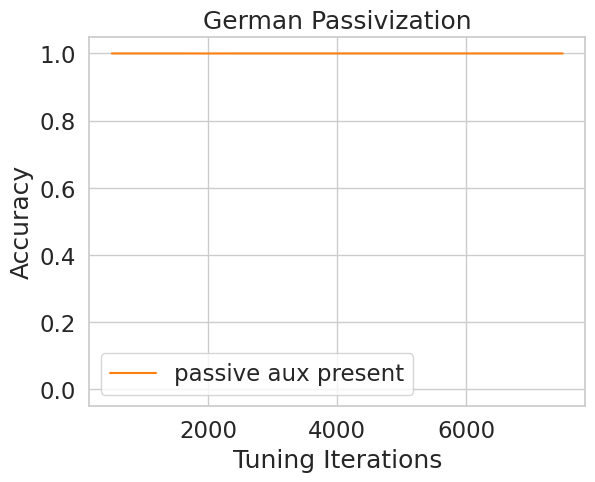}
    \caption{Learning curves displaying alternative accuracy metrics for mT5 on German passivization. We present the proportion of examples for which the model moves the first NP without reinflecting its case (top left), moves the second NP without reinflecting its case (top right), reinflects the tense of the main verb (bottom left), and inserts the passive auxiliary \textit{werden} with the proper inflection.}
    \label{fig:mt5_full_error_analysis}
\end{figure}

\begin{figure}
    \centering
    \includegraphics[width=0.49\linewidth]{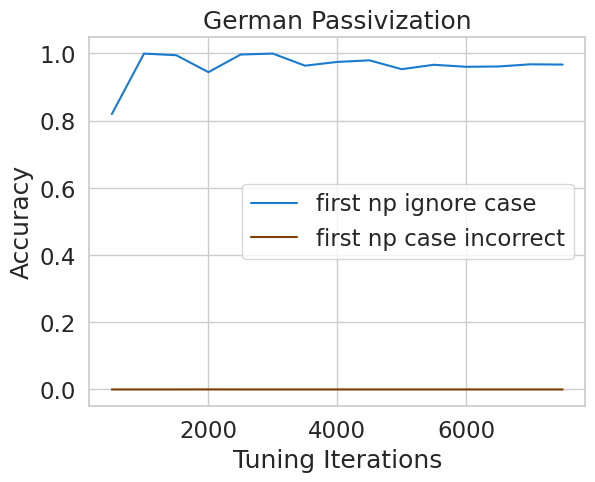}
    \hfill
    \includegraphics[width=0.49\linewidth]{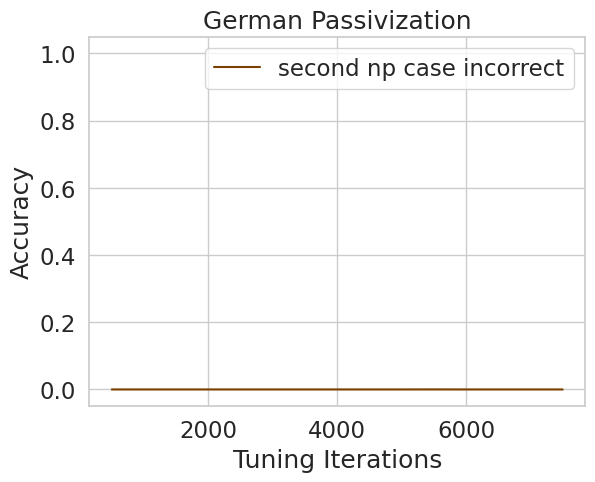}
    \newline
    \includegraphics[width=0.49\linewidth]{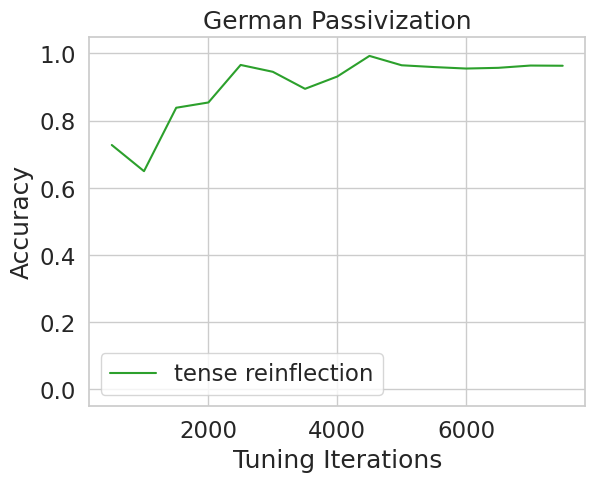}
    \hfill
    \includegraphics[width=0.49\linewidth]{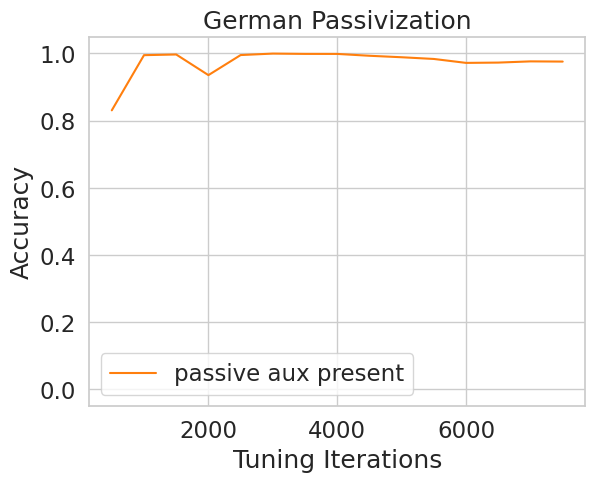}
    \caption{Learning curves displaying alternative accuracy metrics for mBART on German passivization. We present the proportion of examples for which the model moves the first NP without reinflecting its case (top left), moves the second NP without reinflecting its case (top right), reinflects the tense of the main verb (bottom left), and inserts the passive auxiliary \textit{werden} with the proper inflection.}
    \label{fig:mbart_full_error_analysis}
\end{figure}

We also present more detailed analyses of other required operations in passivization: namely, are mT5 and mBART capable of tense reinflection, case reinflection, and auxiliary insertions? And are they capable of this in zero-shot settings? Results for mT5 (Figure~\ref{fig:mt5_full_error_analysis}) and mBART (Figure~\ref{fig:mbart_full_error_analysis}) suggest that both models are generally capable of tense reinflection, case reinflection, and auxiliary insertion in supervised contexts.

\begin{figure}
    \centering
    \includegraphics[width=0.49\linewidth]{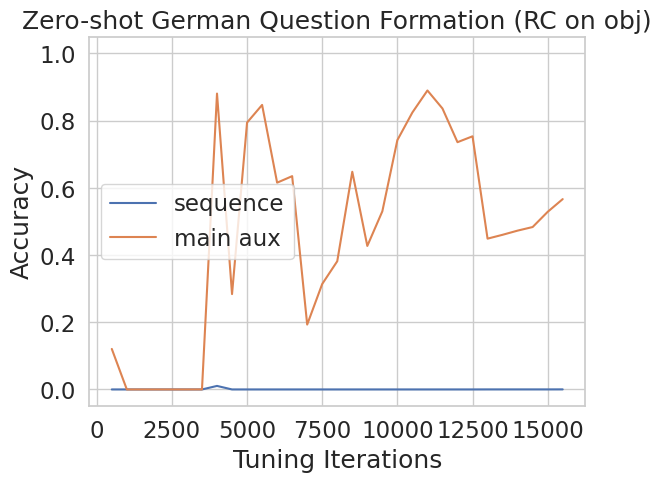}
    \hfill
    \includegraphics[width=0.49\linewidth]{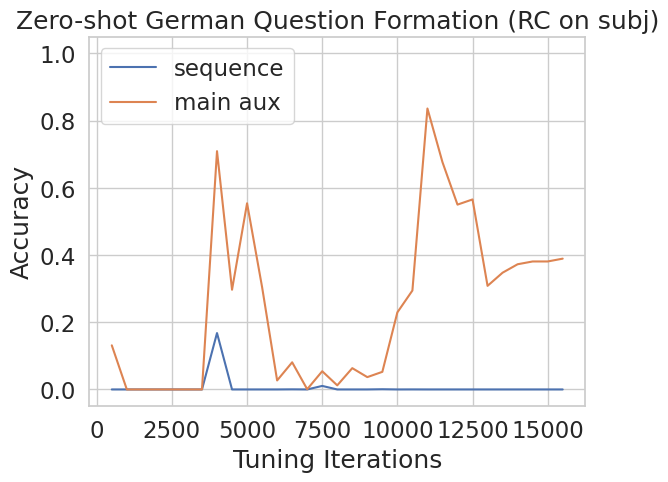}
    \newline
    \includegraphics[width=0.49\linewidth]{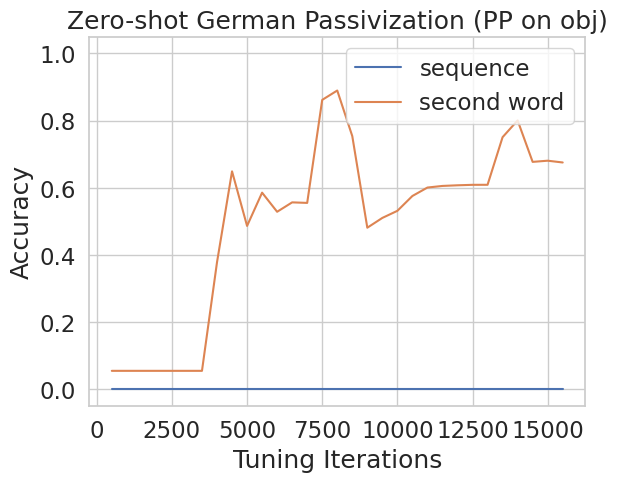}
    \hfill
    \includegraphics[width=0.49\linewidth]{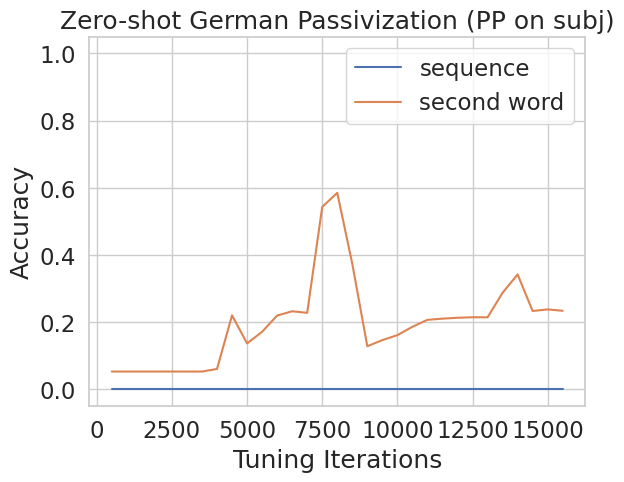}
    \caption{Learning curves for mBART on German transformations after fine-tuning only on English/German identity examples and English transformations. We show accuracies for German question formation with RCs on objects (top left) and RCs on subjects (top right), as well as accuracies for German passivization with PPs on objects (bottom left) and PPs on subjects (bottom right).}
    \label{fig:zeroshot_mbart_de}
\end{figure}

\section{Zero-shot mBART Accuracies}\label{app:zeroshot_mbart}
Here, we present learning curves for mBART on zero-shot cross-lingual syntactic transformations (Figure~\ref{fig:zeroshot_mbart_de}). While mBART is typically able to select the correct auxiliary verb or object noun to move, it never transforms the sequence fully correctly.
\end{document}